# The Effect of Top-Down Attention in Occluded Object Recognition

Zahra Sadeghi

Institute for Research in Fundamental Sciences (IPM)

zahsade@gmail.com

**Abstract**
This study is concerned with the top-down visual processing benefit in the task of occluded object recognition. To this end, a psychophysical experiment is designed and carried out which aimed at investigating the effect of consistency of contextual information on the recognition of objects which are partially occluded. The results demonstrate the facilitative impact of consistent contextual clues on the task of object recognition in presence of occlusion.

**Introduction**
Our experience about each object in the world occurs in conjunction with other objects that tend to be found in similar environments (Xavier Alario, Segui, & Ferrand, 2000). It has been demonstrated that brain processes the gist information of the scene in a quick glance and incorporates this contextual information in order to identify objects in a scene (Oliva, 2005)(Oliva & Torralba, 2007). In particular, it has been shown that object recognition process can be influenced by top-down information and the relevant contextual information can facilitate the perception procedure by providing informative cues about the items that share similar backgrounds(Bar et al., 2006) (Davenport & Potter, 2004).

**Method and Results**
The focus of this study is on the advantage of consistency of top-down information in recognizing objects in difficult situations. This research is concerned with the effect of semantically consistent /inconsistent contextual priming in rapid occluded object recognition. To this end, a psychophysical experiment is conducted in which participants were asked to identify the category of partial objects that were presented on the screen (Figure 1 and Table 1). Before the presence of each object, participants briefly viewed a scene image that could be either consistent or inconsistent with the category of the following occluded objects. According to our hypothesis, the gist information of this image will direct attention to the common objects that tend to be found in similar environments. Hence, top-down attention signal modulates brain and rises a prediction coding about what might be seen next. The objects are chosen from two subcategories of vehicle/non-vehicle items.

In order to evaluate the results, hit rate, miss rate and the response time over the congruent and incongruent pairs of scene and object images in the basic and superordinate modes are measured. In the basic mode, true recognition was considered based on correct category detection, while in the superordinate mode, the confusion between objects from similar superordinate categories of vehicle/non-vehicle was also considered as correct classification. The results confirm that consistent superordinate clues facilitate the process of object recognition. The findings of this study suggest that coupling occluded objects with consistent backgrounds leads to higher accuracy and lower time responses as opposed to inconsistent prior information in occluded object recognition task[1].

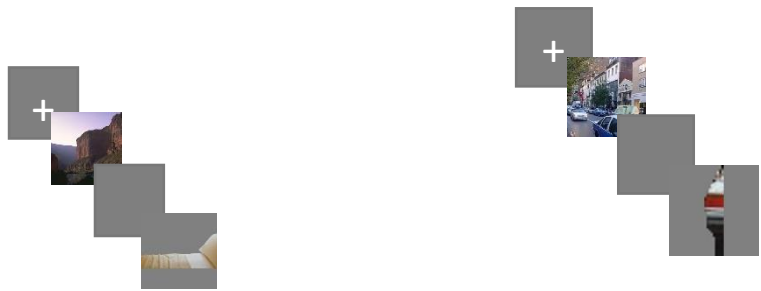

Figure 1. Experiment paradigm. First the fixation pattern is shown on the screen. Next a scene image followed by a partially visible object image appears. The scene image could be either contextually consistent (such as the picture shown in the right side) or inconsistent (such as the picture shown in the left side) with the object.

> % instruction on screen:
> *You will see a scene image followed by an image of an individual occluded object*
>
> *Type the name of the object and then press enter*
>
> *Objects belong to the following categories: 'bed', 'bus', 'car', 'chair'*
>
> % select an object category randomly
>
> % sample an occluded object image from the selected object category
>
> % sample a scene image{with probability of 0.5 select a consistent or inconsistent scene}
>
> % make the screen gray
>
> % show the fixation point

---

[1] The results of this study is presented in the 8'th Iberian Conference on Perception (CIP2019)

```
% show the scene image

% make the screen gray

% show the occluded object image

% make the screen gray

% instruction on screen:
```
*Type the name of the object you just saw, then press enter*

<div align="center">Table 1. Experiment methodology</div>

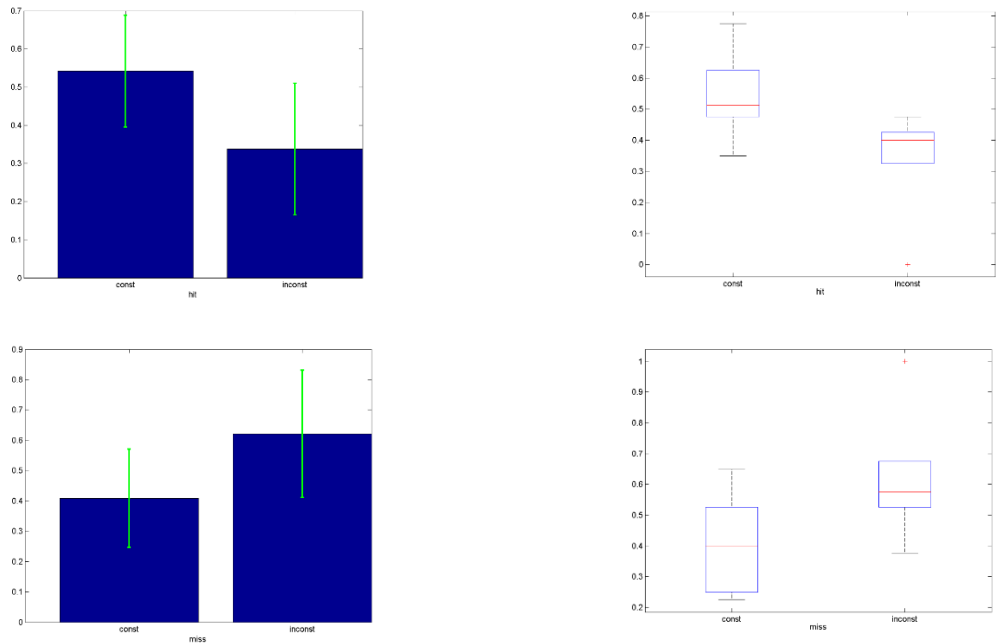

Figure 2. Hit rate and miss rate in occluded object recognition for consistence (const) vs inconsistence (incost) cases in the basic mode.

| Metric   | Hit     | Miss    | RT          |
|----------|---------|---------|-------------|
| p-value  | 0.0027* | 0.0027* | 4.6921e-11* |

Table 2. P-values corresponding to the effectiveness of consistent clues for object recognition in the basic mode. Accuracy of recognition in terms of hit/miss rates and response time (RT) in occluded object recognition are considered. P-values indicate whether the difference in accuracy of recognition between consistent and inconsistent cases are significant. The significant cases are marked by a star.

| Metric | Hit | Miss |
|---|---|---|
| p-value | 0.0027* | 0.0027* |

Table 3. P-values corresponding to the effectiveness of consistent clues from superordinate categories for object recognition in the superordinate mode. Accuracy of recognition in terms of hit/miss rate in occluded object recognition is considered. P-values indicate whether the difference in accuracy of recognition between consistent and inconsistent cases are significant. The significant cases are marked by a star.